\documentclass{article}

\PassOptionsToPackage{numbers}{natbib}

\usepackage[preprint]{neurips}

\usepackage[utf8]{inputenc} 
\usepackage[T1]{fontenc}    
\usepackage{hyperref}       
\usepackage{url}            
\usepackage{booktabs}       
\usepackage{amsfonts}       
\usepackage{nicefrac}       
\usepackage{microtype}      
\usepackage{xcolor}         
\usepackage{adjustbox}
\usepackage{multirow}
\usepackage{subcaption}
\usepackage{colortbl}
\usepackage{xspace}
\usepackage{lipsum}
\usepackage{enumitem}
\usepackage{amsmath}
\usepackage{duckuments}
\usepackage{algpseudocode}  
\usepackage{setspace}       
\usepackage{scrextend}
\usepackage{mathtools}
\usepackage{graphicx}
\usepackage{subcaption}
\usepackage{enumitem}
\usepackage{colortbl}
\usepackage{wrapfig}

\usepackage{soul}
\usepackage{xcolor}
\usepackage{listings}
\usepackage{amssymb}
\usepackage[linesnumbered, boxed, ruled]{algorithm2e}
\usepackage{lineno} 

\definecolor{pinegreen}{rgb}{0.0, 0.47, 0.44}
\definecolor{cornellred}{rgb}{0.7, 0.11, 0.11}
\definecolor{cadmiumgreen}{rgb}{0.0, 0.42, 0.24}
\definecolor{royalblue}{rgb}{0.0, 0.14, 0.4}
\definecolor{spirodiscoball}{rgb}{0.06, 0.75, 0.99}
\definecolor{mylightblue}{rgb}{0.85, 0.90, 0.94}
\definecolor{kaistblue}{RGB}{20,135,200}
\definecolor{auburn}{RGB}{166,38,57}
\definecolor{Gray}{gray}{0.9}
\definecolor{BrickRed}{rgb}{0.6,0,0}
\definecolor{RoyalBlue}{rgb}{0,0,0.8}
\definecolor{Tdgreen}{rgb}{0,0.4,0.7}
\definecolor{darkblue}{rgb}{0,0.08,0.45}
\hypersetup{citecolor=darkblue, linkcolor=darkblue}

\hypersetup{
  urlcolor = cadmiumgreen,
  linkcolor = cornellred,
  citecolor  = kaistblue,
  colorlinks = true,
}

\title{
Tuning-Free Visual Customization via View Iterative Self-Attention Control
}

\newcommand{\sname}{\textit{VisCtrl}\xspace} 
\newcommand{\encoder}{\mathcal{E}}
\newcommand{\decoder}{\mathcal{D}}
\newcommand{\inputimage}{\mathcal{I}}

\newcommand{\model}{\epsilon_\theta}

\newcommand{\conditioner}{\tau_\theta}
\newcommand{\textembedding}{\mathcal{C}}
\newcommand{\textprompt}{\mathcal{P}}

\author{%
    Xiaojie Li$^{1}$, Chenghao Gu$^{2}$, Shuzhao Xie$^{1}$, Yunpeng Bai$^{3}$, Weixiang Zhang$^{1}$, Zhi Wang$^{1}$\thanks{Corresponding author}\\
    $^1$ Tsinghua Shenzhen International Graduate School, Tsinghua University, Shenzhen, China\\
    $^2$ Jiluan Academy, Nanchang University, Nanchang, China\\
    $^3$ Department of Computer Science, The University of Texas at Austin, US\\
    {\small \texttt{\{li-xj23, zhang-wx22\}@mails.tsinghua.edu.cn, wangzhi@sz.tsinghua.edu.cn}}
}

\begin{document}

\maketitle

\begin{abstract}

Fine-Tuning Diffusion Models enable a wide range of personalized generation and editing applications on diverse visual modalities. While Low-Rank Adaptation (LoRA) accelerates the fine-tuning process, it still requires multiple reference images and time-consuming training, which constrains its scalability for large-scale and real-time applications. In this paper, we propose \textit{View Iterative Self-Attention Control (\sname)} to tackle this challenge. Specifically, \sname is a training-free method that injects the appearance and 
structure of a user-specified subject into another subject in the target image, unlike previous approaches that require fine-tuning the model. Initially, we obtain the initial noise for both the reference and target images through DDIM inversion. Then, during the denoising phase, features from the reference image are injected into the target image via the self-attention mechanism. Notably, by iteratively performing this feature injection process, we ensure that the reference image features are gradually integrated into the target image. This approach results in consistent and harmonious editing with only one reference image in a few denoising steps. Moreover, benefiting from our plug-and-play architecture design and the proposed Feature Gradual Sampling strategy for multi-view editing, our method can be easily extended to edit in complex visual domains. Extensive experiments show the efficacy of \sname across a spectrum of tasks, including personalized editing of images, videos, and 3D scenes. 
\end{abstract}

\section{Introduction}\label{sec:intro}

Imagine a world where visual creativity knows no bounds, liberated from the drudgery of manual editing and long waits. In this realm, you can swiftly manipulate diverse visual scenes: seamlessly integrating your beloved cat into any photograph, tailoring landscapes to your liking within VR/AR, or substituting individuals in videos with anyone you choose. This question lies at the heart of a challenging task—\textit{rapidly personalized visual editing} which involves efficiently injecting user-specified visual features (e.g. appearance and structure) into the target visual representation.

The solutions for the personalized visual editing task fall into two paradigms: model-based and attention-based methods. Model-based methods~\cite{brooks2022instructpix2pix,chen2023anydoor,yang2022paint} focus on collecting datasets to fine-tune the entire model, which requires substantial time and computational resources. To avoid the costly process, attention-based methods~\cite{hertz2022prompt,cao2023masactrl,parmar2023zero} have been proposed, with a special focus on manipulating the attention in the UNet of the diffusion model. Prompt-to-Prompt~\cite{hertz2022prompt} can edit images by injecting cross-attention maps during the diffusion process through editing only the textual prompt. MasaCtrl~\cite{cao2023masactrl} utilizes mutual self-attention to achieve non-rigid and consistent image editing by querying correlated local contents and textures from the source image. The editing methods for other visual modalities, such as video and 3D scenes, mostly build upon the aforementioned image editing techniques~\cite{instructnerf2023,ceylan2023pix2video}

However, previous methods still face several challenges in the efficiency of personalized visual editing: \textbf{1)} The prolonged DDIM inversion process causes the intermediate codes to diverge from the original trajectory, leading to unsatisfying image reconstruction~\cite{mokady2022null}. \textbf{2)} The inherent ambiguity and inaccuracy of text often result in significant disparities between the user’s desired content and the generated output~\cite{cao2023masactrl}. Furthermore, even minor adjustments to prompts in most text-to-image models can result in significantly different images~\cite{hertz2022prompt}. \textbf{3)} These methods lack support for other visual representations, hindering their extension to video and 3D scene editing.

\begin{figure}[t]
  \centering
  \includegraphics[width=1.000\textwidth]{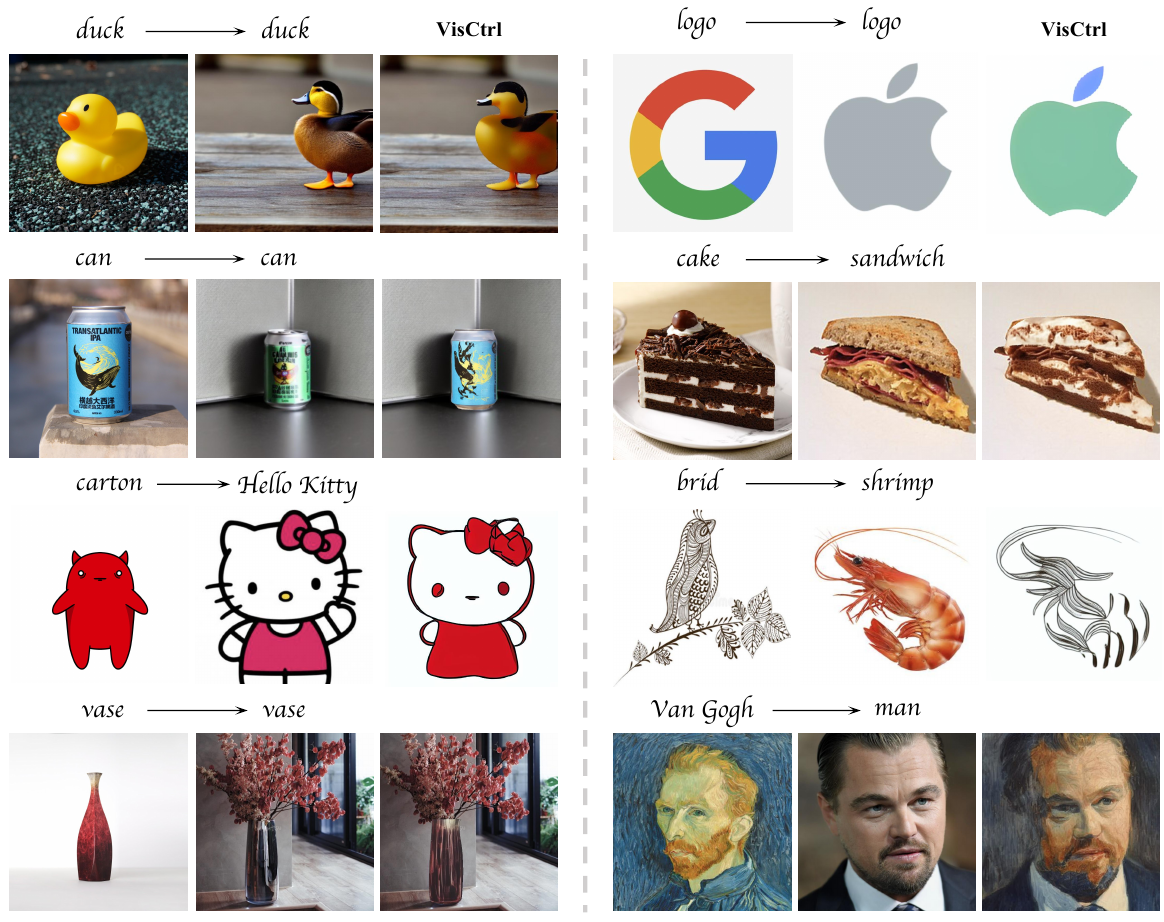}
  \caption{\textbf{\sname{} results span across various object and image domains, showcasing its broad applicability.}  From simple objects (cartoons, logos) to complex subjects (food, humans), the diversity in personalized image editing highlights the versatility and robustness of our framework across different usage scenarios.}
  \label{fig:show_res}
\end{figure}

To tackle these challenges, we propose View Iterative Self-Attention Control (\sname), a simple but effective framework that utilizes self-attention to inject personalized subject features into the target image. Specifically, we firstly obtain the initial noise for both the reference image and the target image through DDIM inversion~\cite{song2021ddim}. Subsequently, during denoising reconstruction, we iteratively inject the features of user-specified subject into the target image using self-attention, while maintaining the overall structure of the target image using cross-attention. Additionally, we propose a Feature Gradually Sampling strategy for complex visual editing, which involves randomly sampling the latent feature from the reference images to achieve multi-view editing. Remarkably, We can generate outstanding results in Figure~\ref{fig:show_res} with few denoising steps using only one reference image without retraining. 

Our method is validated through extensive experiments and shows promise for  extension to other visual personalized tasks. Our contributions are summarized as follows: 
\textbf{1)} We propose a training-free framework for image editing with only one reference image, emphasizing speed and efficiency. 
\textbf{2)} We propose an iterative self-attention control that utilizes the reference image and corresponding textual conditions to govern the editing process. 
\textbf{3)} We propose a Feature Gradually Sampling strategy which effectively extends our framework to other visual domains, such as video and 3D scenes.

\section{Related work}\label{sec:rel}
\begin{figure}[t]
  \centering
    \includegraphics[width=1.000\textwidth]{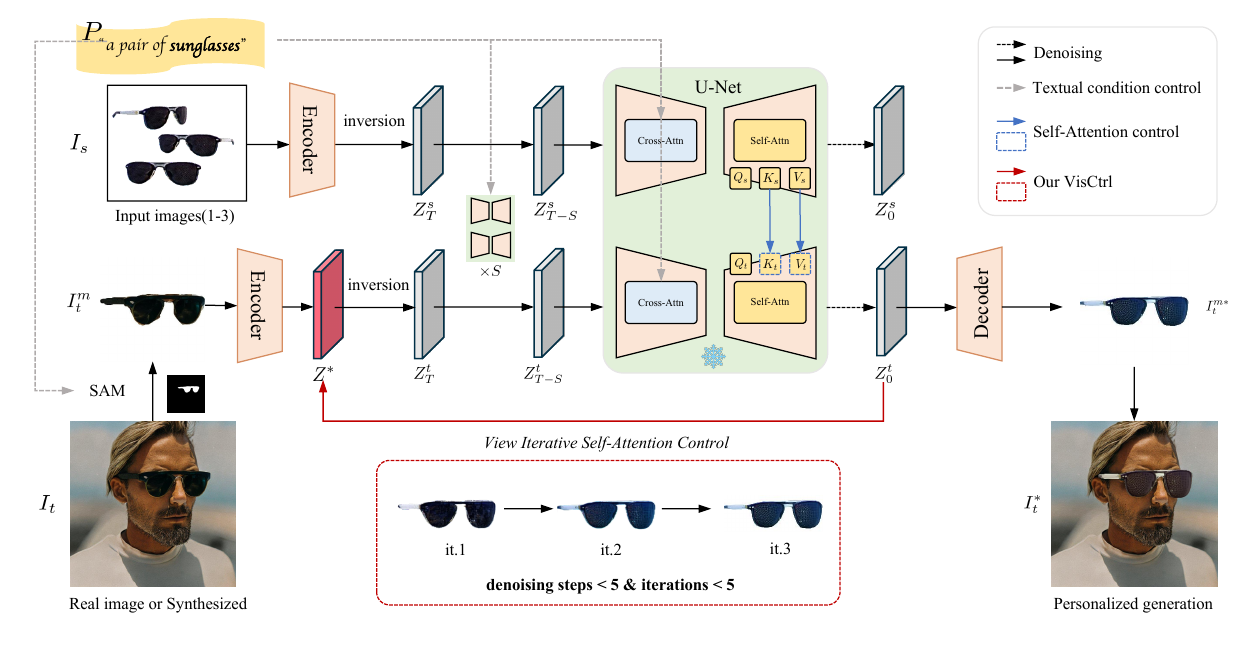}
  \caption{Pipeline of the proposed \sname. Given one or several reference images of a new concept, we first encoder them to the latent space, followed by adding noise and denoising via DDIM~\cite{song2021ddim}. The upper part of the process entails generating the reference image, while the bottom part involves generating the target image. Specifically, during the denoising process, we replace the $K_t, V_t$ of the target image self-attention layer with $K_s, V_s$ from the reference image self-attention layer. Additionally, we update $Z^*$ with $Z_0^t$ iteratively throughout this process. Finally, we decoder $Z_0^t$ to obtain the target image. Please refer to Section~\ref{sec:visctrl} for further details.
  }
  \label{fig:pipeline}
\end{figure}

\label{sec:related-work}
\subsection{Text-guided Visual Generation and editing}

Early image generation methods conditioned on text description mainly based on GANs~\cite{brock2018large,wang2017generative,karras2019style,ye2021improving,tao2022df}, due to their powerful capability of high fidelity image synthesis. Recent advancements in Text-to-Image (T2I) generation have witnessed the scaling up of text-to-image diffusion models~\citep{saharia2022photorealistic, ramesh2022hierarchical, nichol2021glide} through the utilization of billions of image-text pairs~\citep{schuhmann2022laion} and efficient architectures~\citep{ho2020denoising, song2020score, song2020denoising,Peebles2022DiT, rombach2022high}. These models demonstrate remarkable proficiency in synthesizing high-quality, realistic, and diverse images guided by textual input.
Additionally, they have extended their utility to various applications, including image-to-image translation~\citep{meng2021sdedit, bar2022text2live, hertz2022prompt, kawar2022imagic, brooks2022instructpix2pix, mokady2022null,voynov2022sketch}, controllable generation~\citep{zhang2023adding}, and personalization~\citep{gal2022image, ruiz2022dreambooth}. Recent research has explored various extensions and applications of text-to-image (T2I) models. For instance, Tune-A-Video~\citep{wu2023tune} utilizes T2I diffusion models to achieve high-quality video generation. Additionally, leveraging 3D representations such as NeRF~\cite{mildenhall2021nerf} or 3D Gaussian splatting~\cite{kerbl3Dgaussians}, T2I models have been employed for 3D object generation~\cite{poole2022dreamfusion,stable-dreamfusion,zou2023triplane,tang2023dreamgaussian,liu2023zero} and editing~\cite{instructnerf2023,GaussianEditor}. Text-guided image editing has evolved from early GAN-based approaches~\cite{reed2016generative,zhang2018stackgan++,li2019controllable,gal2022stylegan}, which were limited to specific object domains, to more versatile diffusion-based methods~\cite{zhang2023adding,nichol2021glide,feng2023training-free}. 
However, existing diffusion model methods~\cite{nichol2021glide, avrahami2022blended, meng2021sdedit, hertz2022prompt, mokady2022null} often require manual masks for local editing, and struggle with layout preservation.

\subsection{Subject-driven image editing}
Exemplar-guided image editing covers a broad range of applications, and most of the works~\cite{wang2019example,zhou2021cocosnet} can be categorized as exemplar-based image translation tasks, conditioning on various information, such as stylized images~\cite{liu2021adaattn, deng2022stytr2, zhang2022inversion}, layouts~\cite{yang2022reco, li2023gligen, jahn2021high}, skeletons~\cite{li2023gligen}, sketches/edges~\cite{seo2022midms}. With the convenience of stylized images, image style transfer~\cite{liao2017visual,zhang2020cross,tumanyan2022splicing} receives extensive attention, replying to methods to build a dense correspondence between input and reference images, but it cannot deal with local editing and shape editing. To achieve local editing with non-rigid transformation, conditions like bounding boxes and masks are introduced, but require drawing efforts from users, which sometimes are hard to obtain~\cite{yang2022paint,chen2023anydoor,li2023dreamedit}. A recent work~\cite{gu2023photoswap} learns the visual concept of the subject from reference images and then swaps it into the target image using pre-trained diffusion models. However, it requires multiple reference images to learn the corresponding visual concepts that need to fine-tune diffusion model and a significant number of DDIM inversion and denoising steps, which are time-consuming. Our method leverages attention mechanisms to enable personalized editing without the need for additional training while preserving the identity of the original image.

\section{Method}\label{sec:method}
\begin{figure}[t]
  \centering
    \includegraphics[width=1.000\textwidth]{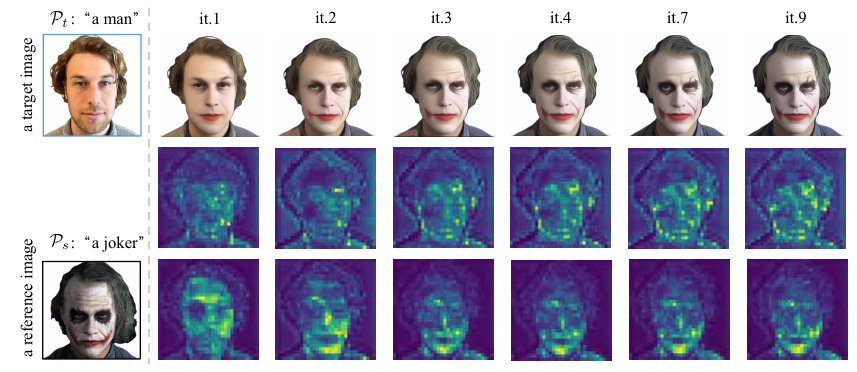}
  \caption{\textbf{Cross-Attention maps under different iterations.} On the left, using the \sname method, the appearance of a reference image with text condition $\textprompt_s$ is inserted into a target image with text condition $\textprompt_t$. On the right are the changes in the target image during the iterations, as well as the changes in the cross-attention computed between its intermediate latent and $\textprompt_s$ and $\textprompt_t$ respectively. Please refer to Section~\ref{sec:2d_editing} for more details.}
  \label{fig:cross_attention}
\end{figure}

In this section, we first provide a short preliminary Section~\ref{sec:preliminary} and then describe our method Section~\ref{sec:visctrl}. An illustration of our method is shown in Figure~\ref{fig:pipeline} and Algorithm~\ref{alg:visctrl}.

\subsection{Preliminary}
\label{sec:preliminary}
\vspace{0.05in}
{\bf Latent Diffusion Models.} Latent Diffusion Model (LDM)~\cite{Rombach_2022_CVPR_stablediffusion} is composed of two main components: an autoencoder and a latent diffusion model. The encoder $\encoder$ from the autoencoder component of the LDMs maps an image $\inputimage$ into a latent code $z_0=\encoder(\inputimage)$ and the decoder reverses the latent code back to the original image as $\decoder(\encoder(\inputimage)) \approx \inputimage$. 
Let $\textembedding=\conditioner(\textprompt)$ be the conditioning mechanism that maps a textual condition $\textprompt$ into a conditional vector for LDMs, the LDM model is updated by the loss:

\begin{equation}
L_{LDM} := \mathbb{E}_{z_0 \sim \encoder(\inputimage), \textprompt, \epsilon \sim \mathcal{N}(0, 1), t \sim \text{U}(1,T) }\Big[ \Vert \epsilon - \model(z_{t},t, \textembedding) \Vert_{2}^{2}\Big] \, 
\label{eq:ldm_loss}
\end{equation}

The denoiser $\model$ is typically a conditional U-Net~\cite{ronneberger2015unet} which predicts the added gaussian noise $\epsilon$ at timestep $t$. 
Text-to-image diffusion models~\cite{saharia2022photorealistic, ramesh2022hierarchical, rombach2022high,nichol2021glide} are trained by Equation~\ref{eq:ldm_loss} with $\model$ that estimates the noise conditioned on the text prompt $\textprompt$. 



{\bf DDIM inversion.} Inversion involves finding an initial noise $z_T$ that reconstructs the input latent code $z_0$ conditioned on $\textprompt$. As our goal is to precisely reconstruct a given image with a reference image, we utilize deterministic DDIM sampling~\cite{song2021ddim}:
\begin{equation}
    z_{t+1} = \sqrt{\bar{\alpha}_{t+1}}f_\theta(z_t,t,\textembedding) + \sqrt{1-\bar{\alpha}_{t+1}} \model(z_t,t,\textembedding)
    \label{eq:ddim}
\end{equation}
where $\bar{\alpha}_{t+1}$ is noise scaling factor defined in DDIM~\cite{song2021ddim} and $f_\theta(z_t,t,\textembedding)$ predicts the final denoised latent code $z_0$ as $f_\theta(z_t,t,\textembedding) = \Big[z_t - \sqrt{1-\bar{\alpha}_t} \model(z_t,t,\textembedding) \Big] / {\sqrt{\bar{\alpha}_t}}$.

{\bf Attention Mechanism in LDM.} The U-Net in the Diffusion model, consists of a series of basic blocks, and each basic block contains a residual block~\cite{he2016deep}, a self-attention module, and a cross-attention~\cite{vaswani2017attention} module. The attention mechanism can be formulated as follows:
\begin{equation}
    \label{eq:attention}
    \text{Attention}\mathcal(Q_t, K, V)= softmax(\frac{Q_{t}K^T}{\sqrt{d}})V,
\end{equation}
where $d$ represents the latent dimension, and $Q$ denotes the query features projected from spatial features, while $K$ and $V$ signify the key and value features projected from the spatial features in self-attention layers or the textual embedding in cross-attention layers. The attention map is $\mathcal{A}_t=softmax(Q_t \cdot K^T / \sqrt{d})$ which is the first component of Equation~\ref{eq:attention}.

\subsection{VisCtrl: View iterative Self-Attention Control}
\label{sec:visctrl}

In this section, we introduce \emph{View iterative self-attention Control} ~(\sname) for Tuning-Free personalized visual editing. The overall architecture of the proposed pipeline to perform synthesis and editing is shown in Figure~\ref{fig:pipeline}, and the algorithm is summarized in Algorithm~\ref{alg:visctrl}. Our goal is to inject the features of the personalized subject in reference images $\{I_s\}_1^N$ (typically 1-3) into another subject $I_t^{m}$ in a given target image $I_t$. Firstly, we use SAM~\cite{kirillov2023segany} to segment the target subject $I_t^{m}$ based on the target text prompt $\textprompt_t$. Then, we obtain the initial noise $Z_T^s$ for the reference images and the initial noise $Z_T^t$ for the target subject through DDIM inversion~\cite{song2021ddim}, which are used for the reconstruction of images. Next, through the U-Net, we obtain the features $K$ and $V$ of the images. Finally, during the target image reconstruction process conditioned on the noise $Z_T^t$ and the target text prompt $\textprompt_t$, the target subject features ($K_t$, $V_t$) are replaced with the reference image features ($K_s$, $V_s$) obtained during the reference image reconstruction process. Hence, we can seamlessly integrate the generated subject back into the target image in a harmonious manner.

\begin{wrapfigure}{r}{0.5\textwidth}
\vspace{-5mm}
\hspace{2mm}
{
\begin{algorithm}[H]
\SetAlgoLined
\textbf{Input:}The reference images $\{I_s\}_1^N$ and corresponding prompt $\textprompt_s$, a target image$I_t$ and corresponding prompt $\textprompt_t$.\\

\textbf{Output:} Edited latent map $z^t_0$. \\
$\{z^s_T\}_1^N\leftarrow \text{DDIMInversion}(\encoder(\{I_s\}_1^N),\textprompt_s)$;\\
$z^*=\encoder(I_t)$;\\
$z^t_T\leftarrow \text{DDIMInversion}(z^*,\textprompt_t)$;\\
\For{$n=N,N-1,\ldots,1$}{
$z^t_T \leftarrow \alpha * \text{DDIMInversion}(z^*,\textprompt_t) + (1-\alpha)*z^t_T$;\\
$z^s_T=\text{DataSampler}(\{z^s_T\}_1^N)$;\\
 \For{$t=T,T-1,\ldots,1$}{
    $\epsilon_s, \{Q_s, K_s, V_s\}\leftarrow \epsilon_\theta(z^s_t, P_s, t)$;  \\ 
    $z^s_{t-1} \leftarrow \text{DDIMSampler}(z^{s}_{t}, \epsilon_s)$;\\
    $\{Q_t, K_t, V_t\} \leftarrow \epsilon_\theta(z_t,P,t)$;\\
    $\{Q_t^*, K_t^*, V_t^*\} \hspace{-1mm}\leftarrow \text{Edit}(\{Q_t,K_t,V_t\}, \{Q_s, K_s, V_s\})$;\\
    $\epsilon_t = \epsilon_\theta(z_t, P, t; \{Q_t^*, K_t^*, V_t^*\})$;\\
    $z^t_{t-1} \leftarrow \text{DDIMSampler}(z_t^t, \epsilon_t)$;\\
 }
 $z^* =z^t_0$;\\
}
 \textbf{Return} $z^t_0$ \\
 \caption{View Iterative Self-Attention Control}
\label{alg:visctrl}
\end{algorithm}
}
\end{wrapfigure}

As shown in Figure~\ref{fig:pipeline}, the architecture includes the reference image branch (top) and the target image branch (bottom), both branches perform inversion and denoising, but the denoising process will be different. Specifically, the reference image branch provides personalized subject features through self-attention. Then, in the target image branch, we assemble the inputs for the self-attention by \textbf{1)} keeping the current Query features $Q_t$ unchanged, and \textbf{2)} obtaining the Key and Value features $K_s$ and $V_s$ from the self-attention layer in the reference image branch. \textbf{3)} 
Continuously perform the denoising process described above to obtain $Z_0^t$. \textbf{4)} Finally, utilize $Z_0^t$ as a replacement for $Z^*$, followed by an inversion process and iterate through steps (1), (2), and (3) for $N$ iterations to gradually inject the feature of reference images into the target image. We initialize $Z^*$ as $\encoder(I_t)$. During the iterative process, $Z^*$ is updated according to the following formulation:
\begin{equation}
    \label{eq:z_star}
    Z^*_{(n+1)} = Z^t_{0(n)}, \quad 1 < n < N
\end{equation}
where $n$ denotes the iteration number, $N$ is the total number of iterations. It is noteworthy that significant improvement can be achieved within 5 iterations. Each iteration involves an inversion process and denoising process, both of which do not exceed 5 steps. Remarkably, We can control the level of the appearance and structure of reference images into target image with proper starting denoising step $S$ and layer $L$ for editing, please refer to Figure~\ref{fig:diff_step}. Thus the Edit function in Algorithm~\ref{alg:visctrl} can be formulated as follows:

\begin{equation}
    \label{eq:visctrl_edit}
    \text{Edit} := \left \{
    \begin{aligned}
        & \{Q_t, K_s, V_s\},\quad \text{if}\quad t > S \quad \text{and} \quad l > L, \\
        & \{Q_t, K_t, V_t\},\quad \text{otherwise},
    \end{aligned}
    \right.
\end{equation}

where $S$ and $L$ are the time step and layer index to start \sname, respectively.

\subsection{Feature Gradually Sampling strategy for multi-view editing}
\label{sec:gfs}

When applying the \sname method to complex visual domains where the target content is distributed across multiple views, such as video editing and 3D editing, we encounter two key challenges: \textbf{1) Limited usability of single reference Image:} In complex scenarios with multiple perspectives, relying on single reference image often leads to blurring due to significant changes between different views. This occurs because retrieving insufficient useful information from a single reference image can cause the target image to lose its original structure during the iterative process. Once the structure is compromised, it becomes difficult to restore, as the missing structure is no longer present in the query of the target image, please refer to Figure ~\ref{fig:gfs_fusion}. \textbf{2) Consistent injection from multiple reference images:} When incorporating multiple reference images, it's crucial to ensure that the injection of information from these images is consistent. Drastic variations can lead to jitter in video and artifacts in 3D scenes.

Therefore, we propose the Feature Gradual Sampling strategy (FGS) for multi-view editing, which involves randomly sampling the data from the reference images to allow the target image to perceive as much useful information as possible. Additionally, to mitigate forgetting, we will let $z$ with weighted updates during the iterative process. $Z^t_{T(n+1)}$  is updated according to the following formulation:
\begin{equation}
    \label{eq:z_t}
    Z^t_{T(n+1)} = \alpha * \mathcal{F}(Z^*_{(n)}, \textprompt_t) + (1-\alpha)*Z^t_{T(n+1)}, \quad 1 < n < N
\end{equation}
where $n$ denotes the iteration number,  $\mathcal{F}$ represents the process of target branch DDIM inversion, obtaining the initial noise using $Z^*_{(n)}$ under the condition of $\textprompt_t$. The parameter $\alpha$ denotes the sampling coefficient, which controls the degree of feature injection. A smaller $\alpha$ results in more gradual feature changes.

\section{Experiments}

\begin{figure}[t]
  \centering
    \includegraphics[width=1.000\textwidth]{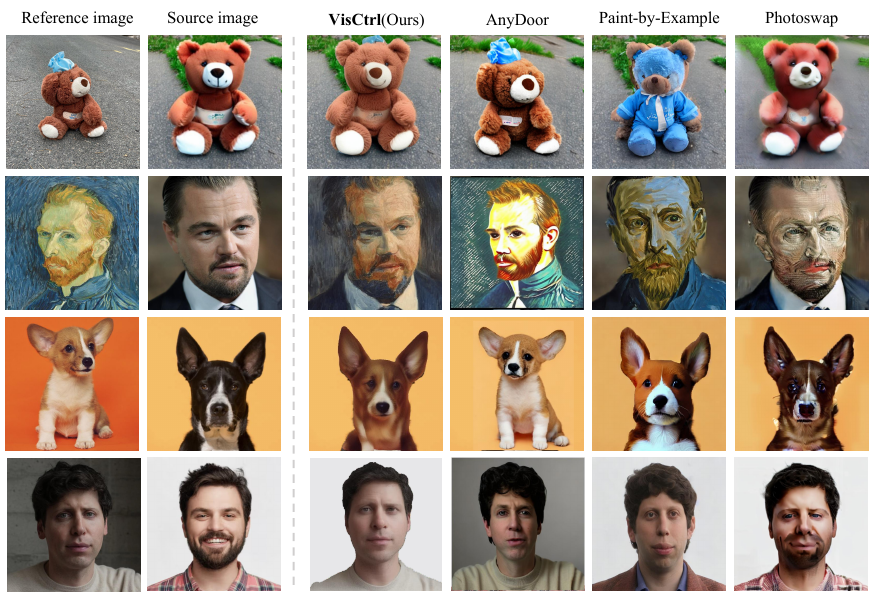}
  \caption{\textbf{Results of different methods on personalized image editing.} Our proposed \sname method yields compelling results across various object and image domains, showcasing its broad applicability. From left to right: the reference image and the source image with their respective prompts, editing results with the proposed \sname{} method, and Other Exemplar-guided Image Editing results with existing methods AnyDoor~\cite{chen2023anydoor}, Paint by Example~\cite{yang2022paint}, and Photoswap~\cite{gu2023photoswap}. Please refer to Section~\ref{sec:comparison} for more details.
  }
  \label{fig:mutil_res}
\end{figure}

Our \sname can be used to edit images, videos, and 3D scenes. We validate the effectiveness of FGS  and demonstrate that \sname can control the degree of subject personalization, including its shape and appearance, please refer to Appendix~\ref{appen:ablation}. We showcase the capabilities of our method in various experiments, please refer
to Appendix~\ref{appen:imp_details}.

\subsection{Personalized Subject Editing in images}
\label{sec:2d_editing}

\begin{table*}[t]
    \caption{
    \textbf{Comparison to prior exemplar-guided image editing methods.} We compare our method with several prior Exemplar-guided Image Editing approaches across three distinct tasks. The initial two editing tasks (dog $\rightarrow$ dog, teddy bear $\rightarrow$ teddy bear) are assessed using CLIP-I, BG LPIPS, and SSIM. Definitions and details of these metrics can be found in the Appendix~\ref{appen:metrics}. Specifically, we contrast the generated images with both the reference image and the source image, resulting in two CLIP-I scores. In the CLIP-I column, the left value denotes the score between the reference image and the generated image, while the right represents the score between the source image and the generated image. For the remaining task (man $\rightarrow$ van gogh), only CLIP-I and SSIM metrics are utilized, as background reconstruction is deemed irrelevant.
    }
    \centering
    \resizebox{\textwidth}{!}{
    \begin{tabular}{l ccc ccc cc}
        \toprule 
        \multirow{3}{*}{\textbf{Method}} 
        & \multicolumn{3}{c}{\textbf{dog $\rightarrow$ dog} }
        & \multicolumn{3}{c}{\textbf{teddy bear $\rightarrow$ teddy bear} }
        & \multicolumn{2}{c}{\textbf{man $\rightarrow$ van gogh} } \\

        \cmidrule(lr){2-4} \cmidrule(lr){5-7} \cmidrule(lr){8-9}

        & \multirow{2}{*}{\shortstack[c]{CLIP-I ($\uparrow$) }}  
        & \multirow{2}{*}{\shortstack[c]{BG\\ LPIPS ($\downarrow$) }} 
        & \multirow{2}{*}{\shortstack[c]{SSIM($\uparrow$) }} 
        
        & \multirow{2}{*}{\shortstack[c]{CLIP-I ($\uparrow$) }}  
        & \multirow{2}{*}{\shortstack[c]{BG\\ LPIPS ($\downarrow$) }}
        & \multirow{2}{*}{\shortstack[c]{SSIM($\uparrow$) }}

        & \multirow{2}{*}{\shortstack[c]{CLIP-I ($\uparrow$) }}
        & \multirow{2}{*}{\shortstack[c]{SSIM($\uparrow$) }}

        \\ \\
        \cmidrule(lr){1-9}

        AnyDoor \cite{chen2023anydoor}
        & \textbf{79.9\%} / 75.3\% & 0.379 & 0.580 
        & 70.2\% / 80.1\% & 0.378 & 0.546
        & 59.6\% / 48.8\% & 0.289 \\

        Paint-by-Example \cite{yang2022paint}
        & 75.6\% / 75.5\% & 0.287 & 0.674
        & \textbf{76.4\%} / 75.1\% & 0.388 & 0.601
        & 64.5\% / 41.1\% & 0.522 
        \\

        Photoswap \cite{gu2023photoswap} 
        & 69.8\% / \textbf{80.8\%} & 0.225 & 0.768 
        & 62.6\% / 78.2\% & 0.228 & 0.640
        & 47.7\% / 51.2\% & 0.635 
        \\
        
        \textbf{\sname(ours) }
        & 76.7\% / 71.9\%  & \textbf{0.211} & \textbf{0.822 }
        & 72.8\% / \textbf{85.7\%} & \textbf{0.205} & \textbf{0.838}
        & \textbf{72.1\%}/ \textbf{69.1\%} & \textbf{0.746}
        \\
        \bottomrule 
    \end{tabular}
    }
    \label{tab:cmp_baselines}
\end{table*}

Figure~\ref{fig:show_res} showcases the effectiveness of \sname for personalized subject editing in images. Our approach excels at preserving crucial aspects such as spatial layout, geometry, and the pose of the original subject while seamlessly introducing a reference subject into the source image. Our method can not only achieve personalized injection of similar subject (e.g. duck to personalized duck, vase to personalized vase) but also enable editing between different subject (e.g. injecting cake features into a sandwich, incorporating Van Gogh's art style into a portrait).

To demonstrate the effectiveness of our feature injecting method, we examined the changes in the generated images and the corresponding cross-attention maps with different prompts under different iterations. As shown in Figure~\ref{fig:cross_attention}, it can be seen that with only 4 iterations, the quality of the generated images can rival that of 9 iterations. As the iterations progress, the features from the reference image 'joker' gradually become richer (e.g. the black eye circles in the second iteration, the wrinkles on the forehead in the third iteration). We compute the cross-attention map related to $\textprompt_t$ and the latent of the target branch (Figure~\ref{fig:pipeline} below) by using Equation ~\ref{eq:attention}, where the features about "joker" continue to manifest, as shown in the middle row. Similarly, we compute the cross-attention map related to $\textprompt_s$ and the latent of the reference branch (Figure~\ref{fig:pipeline} above), where the features about "man" gradually diminish, as shown in the bottom row. In Figure~\ref{fig:vis_self_attention}, we also observe the changes in self-attention during different iterations of the generation process.

\subsection{Comparison with Baseline Methods}\label{sec:comparison}
We compared our method with several baselines for personalized image editing. Please refer
to Appendix~\ref{appen:evaluation} for more details.

In Figure~\ref{fig:mutil_res}, we present a comparative analysis between our approach and the baselines. AnyDoor generated images exhibit favorable features related to subject from reference images, albeit with structural degradation of the source image. Paint-by-Example produces high-quality results but fails to inject subject-related features and adequately preserve the layout structure of the source image. Although Photoswap retains both subject features and the layout structure of the source image, it suffers from inferior generation quality. Our method far surpass those baselines, effectively balancing the preservation of the source image's layout structure and background while incorporating more features from the reference image. 

In Table~\ref{tab:cmp_baselines}, we conduct a comparative analysis between our method and the baselines, revealing a consistent trend. AnyDoor exhibits the highest BG LPIPS score, indicating significant variations in the background of source images. Paint-by-Example generally achieves lower CLIP-I score, suggesting substantial disparities between the generated image and both source and reference image. Our method achieves The first and second highest CLIP-I score, striking a balance between incorporating the appearance features from the reference image and preserving the structural characteristics of the source image. This is evidenced by the lowest BG LPIPS score and the highest SSIM score.

\begin{figure}[!th]
  \centering
    \includegraphics[width=1.000\textwidth]{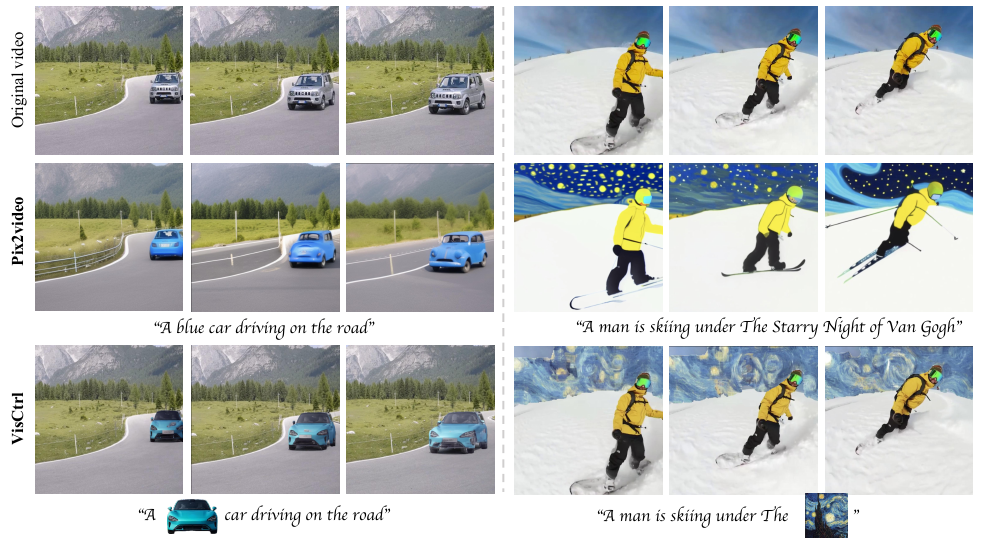}
  \caption{\textbf{Results of different methods on personalized video editing.} We edit the foreground subject and background of various videos using different methods. Compare to baseline, Our method not only generates content that is more similar to the reference image but also maintains the continuity of the edited regions across different frames.
  }
  \label{fig:video_baseline}
\end{figure}

\subsection{Personalized Subject editing in complex visual domains}

Thanks to the following characteristics of our method \sname, our approach can be easily adapted to other complex visual personalized editing tasks: \textbf{1)} The plug-and-play architecture allows direct usage on any method that utilizs Stable-diffusion. \textbf{2)} The distinguishing attribute of our method, Training-Free, is its capability to complete single-image editing within just a few denoising steps without fine-tune.
\textbf{3)} The Feature Gradually Sampling strategy for Multi-view editing (Section~\ref{sec:gfs}) enables consistent editing across multiple views. We conducted a spectrum of experiments in complex visual scenarios, validating the scalability of our method.

{\bf Video editing.} We adopt Pix2video~\cite{ceylan2023pix2video} as our baseline, which utilizes a 2D diffusion model driven by text to achieve image editing. In the task of video editing, we use a single image as a reference subject, and  insert its feature into corresponding subject in each frame of the video. As illustrated in Figure~\ref{fig:video_baseline}, our approach edits the content in the video to be most similar to the reference subject, while effectively controlling the influence on other content outside the editing region. Moreover, as shown in Table~\ref{tab:video_table}, our method achieves the best scores in both CLIP Directional Similarity and LPIPS, indicating that our approach not only preserves the layout of the target image but also effectively achieves personalized editing of the video scene.

{\bf 3D scene editing.}\label{sec:3d_editing} Our method extends upon AnyDoor~\cite{chen2023anydoor} by introducing the VisCtrl module (see more details in Appendix~\ref{appen:3d_detail}), enabling to inject the features of the reference images into the target subject in the 3D scene. Moreover, leveraging the FGS enhances the performance of 2D image editing methods in 3D scene editing. As observed in Figure~\ref{fig:3d_baseline}, Instruct-NeRF2NeRF (IN2N) generated sunglasses exhibit missing structures and even affect irrelevant backgrounds (as shown in the red circles in the figure). The sunglasses generated by AnyDoor differ significantly in appearance and shape (as shown in the blue circles in the figure) from the reference image. 
The noise in the sunglasses generated by AnyDoor is due to the inconsistent editing between different views. These inconsistent edits make it difficult for the 3DGS~\cite{kerbl3Dgaussians} to converge. Our method alleviates this issue by ensuring more consistent editing (as shown in the green circles in the figure).
\sname improves the subject similarity and structural continuity. Quantitative indications in Table ~\ref{tab:3d_table} also clearly demonstrate the significant improvement in effectiveness brought about by the incorporation of the \sname module.

\begin{table}[t]
\centering
\vspace{-0.1in}
\caption{\textbf{Comparison to prior complex visual editing methods}. We individually assess the quantitative metrics of \sname{} in both video editing and 3D scenes, comparing them against other baseline methods.
}\label{tab:3d_video_table}

\begin{subtable}[t]{.55\linewidth}
{
\caption{\textbf{Video editing}. Quantitative comparison of video editing. Our method, \sname{}, is compared with Pix2video across two video scenarios: background editing (e.g. sky) and foreground subject manipulation (e.g. car). \sname{} outperforms on par with existing method across almost metrics.
}\label{tab:video_table}
\resizebox{\textwidth}{!}{
        \begin{tabular}{l ccc ccc}
        \toprule 
        \multirow{3}{*}{\textbf{Method}} 
        & \multicolumn{3}{c}{\textbf{sky $\rightarrow$ sky} }
        & \multicolumn{3}{c}{\textbf{car  $\rightarrow$ car} }\\
        
        \cmidrule(lr){2-4} \cmidrule(lr){5-7}

        & \multirow{2}{*}{\shortstack[c]{CLIP Directional \\Similarity($\uparrow$) }}  
        & \multirow{2}{*}{\shortstack[c]{CLIP-I($\uparrow$) }} 
        & \multirow{2}{*}{\shortstack[c]{LPIPS($\downarrow$) }}
        
        & \multirow{2}{*}{\shortstack[c]{CLIP Directional \\Similarity($\uparrow$) }}  
        & \multirow{2}{*}{\shortstack[c]{CLIP-I($\uparrow$) }} 
        & \multirow{2}{*}{\shortstack[c]{LPIPS($\downarrow$) }}
        \\ \\
        \cmidrule(lr){1-7}

        Pix2video~\cite{ceylan2023pix2video}
        & 0.136 & \textbf{84.3\%} & 0.509 
        & 0.087 & 76.2\% & 0.392 \\
        
        \textbf{\sname(ours)} 
        & \textbf{0.226}  & 82.2\% & \textbf{0.195}
        & \textbf{0.090} & \textbf{77.9\%} & \textbf{0.044}\\
        \bottomrule 
    \end{tabular}
}
}
\end{subtable}
\hfill
\begin{subtable}[t]{.4\linewidth}
{
\caption{\textbf{3D scene editing}. Quantitative comparison of on 3D scene editing. \sname can achieve plug and play. After using \sname,  the capabilities of Anydoor have been significantly improved on 3D scenes editing, which be marked red in the table.
}\label{tab:3d_table}
\resizebox{\textwidth}{!}{
        \begin{tabular}{l ccc}
        \toprule 
        \multirow{2}{*}{\textbf{Method}} 

        & \multirow{2}{*}{\shortstack[c]{CLIP Directional \\Similarity($\uparrow$)}}  
        & \multirow{2}{*}{\shortstack[c]{CLIP-I($\uparrow$) }} 
        & \multirow{2}{*}{\shortstack[c]{LPIPS($\downarrow$)}}
        \\ \\
        \midrule
         IN2N~\cite{brooks2022instructpix2pix} & \textbf{0.210} & 79.0\% & \textbf{0.401}\\
         AnyDoor~\cite{chen2023anydoor} & 0.180 & 76.3\% & 0.529\\
         AnyDoor+\textbf{\sname} & 0.189( {\color{red}+5\%}) & \textbf{79.9\%}( {\color{red}+4.7\%}) & 0.452({\color{red}+14.5\%})\\
         \bottomrule
        \end{tabular}
}
}
\end{subtable}
\end{table}

\section{Conclusion}\label{sec:con}
In this paper, we propose View Iterative Self-Attention Control (\sname), a simple but effective framework designed for personalized visual editing. \sname is capable of injecting features between images using the self-attention mechanism without fine-tuning the model. Furthermore, we propose a Feature Gradually Sampling strategy to adapt \sname to complex visual applications such as video editing and 3D scene editing. We demonstrate the effectiveness of our method in exemplar-guided visual editing, including images, videos, and real 3D scenes, outperforming previous methods both quantitatively and qualitatively.

{\bf Limitations.}\label{sec:limi} Since we use pre-trained diffusion models, there are instances where the results are imperfect due to the inherent limitations of these models. Additionally, our method relies on masks to specify the objects or regions to be edited, and incorrect masks can lead to disharmonious image editing results. Please refer to Appendix~\ref{append:failure_cases} for further details.

{\bf Broader impacts.}\label{sec:broader_imp} Our research introduces a comprehensive visual editing framework that encompasses various modalities, including 2D images, videos, and 3D scenes. While it is important to acknowledge that our framework might be potentially misused to create fake content, this concern is inherent to visual editing techniques as a whole. Furthermore, our method relies on generative priors derived from diffusion models, which may inadvertently contain biases due to the auto-filtering process applied to the vast training dataset. However, \sname has been meticulously designed to mitigate bias within the diffusion model. Please refer to Appendix~\ref{appd:ethics} for further details.

{
\bibliographystyle{unsrtnat}  
\bibliography{main}
}
\newpage

\newpage
\appendix
\onecolumn
\begin{center}
{\bf {\LARGE Appendix}}
\end{center}

\section{Implementation Details}\label{appen:imp_details}

We demonstrate our method in various experiments using Stable Diffusion v1.5~\cite{Rombach_2022_CVPR_stablediffusion}. The segmentation model utilized in the experiment is the LangSAM segmentation algorithm, which is built upon SAM~\cite{kirillov2023segany}, and the GroundingDINO~\cite{liu2023grounding} detection model. All of our experiments were performed using a single NVIDIA V100 GPU.

\begin{figure}[!th]
  \centering
    \includegraphics[width=1.000\textwidth]{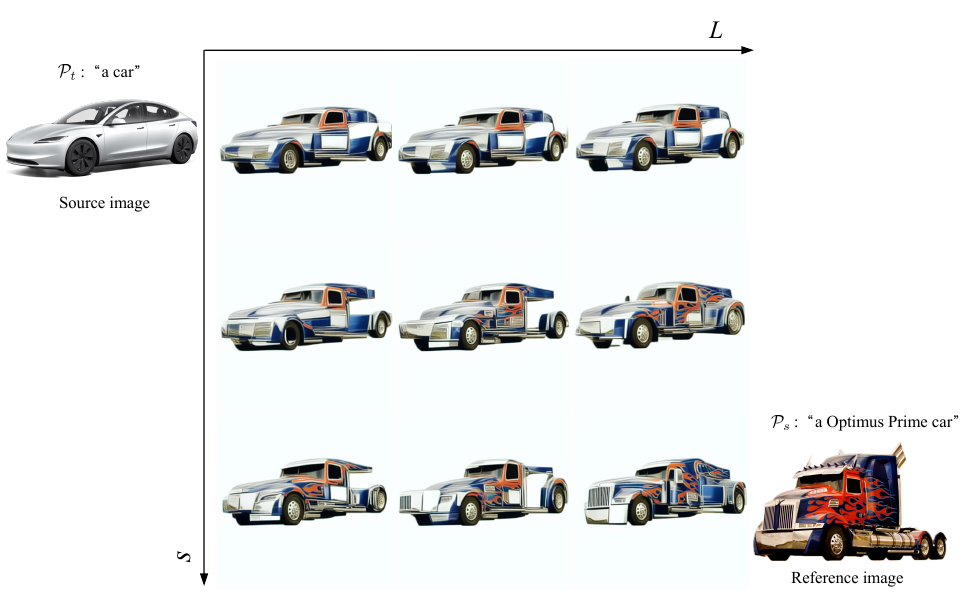}
  \caption{\textbf{Results at different injecting layers and denoising steps.} The top left corner shows the source image and the corresponding text prompt $\textprompt_t$. The bottom right corner displays the reference image and the corresponding text prompt $\textprompt_s$. The middle section presents the generated results with different combinations of the time step  $S$ and the layer index $L$, with the values gradually decreasing in the direction indicated by the arrows.
  }
  \label{fig:shape_ctrl}
\end{figure}

\subsection{2D image personalized editing} 
AnyDoor~\cite{chen2023anydoor} and Paint-by-Example~\cite{yang2022paint} are model-based approaches that require extensive fine-tuning with large datasets. In our experiment, we utilized the default models and parameters as described in their respective papers. Given a source image and mask, the reference image is inserted into the corresponding mask region. Photoswap~\cite{gu2023photoswap} and \sname are attention-based methods that manipulate the attention in UNet to edit images. However, unlike Photoswap, which requires Dreambooth~\cite{ruiz2022dreambooth} to learn new concepts from reference images, \sname does not need any additional training or learning. Since \sname utilizes only one reference image in our experiments, for fairness, we also used a single image for learning new concepts in Photoswap. We set the Dreambooth training steps 1000 for each image in Photoswap, while keeping other parameters at their defaults.

For our method, We set both the noise addition and denoising steps to $T=5$, with classifier-free guidance set to $\omega=6$, and the number of iterations set to $N=5$. Initially, we utilized DDIM Inversion~\cite{song2021ddim} to transform both the reference and target images into initial noise, and then denoising and iteration until convergence. Setting the number of steps higher injects and generates more details, but should not be excessively large to avoid introducing significant biases from DDIM Inversion. In general, during the initial iteration, a higher number of steps can be set to capture more detail, while the denoising steps remain at 5 for subsequent iterations. Our algorithm is highly efficient, typically converging to satisfactory image results within three iterations. It's important to note that in 2D image experiments, only one reference image was used.

\subsection{Video personalized editing} For video editing, we apply our method to edit videos frame by frame. Since few video editing methods support the input of reference images, we compare our model with other text-driven tuning-free video editing models, such as Pix2Video, which can represent common video editing methods. For our work, a reference image is provided to edit each frame of the original video, aiming to achieve the overall editing effect. For the Pix2Video model, we obtain the text description of the reference image and use it as the textual input to achieve the video's editing effect. We set classifier-free guidance $\omega=3.5$, and DDIM steps $T=50$ for Pix2Video. Since Pix2Video does not support the input of reference images, we do not overly discuss the similarity between the editing result and the reference image. Instead, we focus more on the temporal consistency of the edited video and the preservation of the background. 

\subsection{3D Scenes personalized editing} \label{appen:3d_detail}
\begin{figure}[t]
  \centering
    \includegraphics[width=1.000\textwidth]{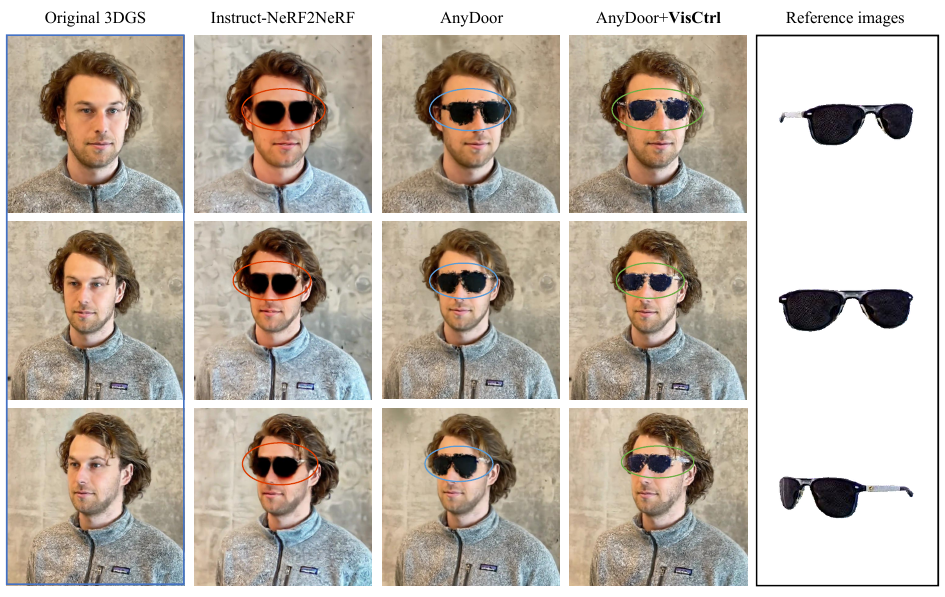}
  \caption{\textbf{Results of different methods on personalized 3D scene editing.} The image on the leftmost is a rendering from the original 3DGS. The image on the rightmost is the reference image used to edit the 3D scene. The images in the middle are rendered from the same viewpoint as the original 3DGS after editing the 3D scene using different methods. We analyze the results of these methods in Section ~\ref{sec:3d_editing}.
  }
  \label{fig:3d_baseline}
\end{figure}

For text-based 3D editing scenes, we use Instruct-NeRF2NeRF as one of our baselines~\citep{instructnerf2023}. We first pretrain 3DGS~\cite{kerbl3Dgaussians} using the \emph{splatfacto} method~\citep{igs2gs} from NeRFStudio~\citep{tancik2023nerfstudio}, training it for 30,000 steps in 10 minutes on an NVIDIA Tesla V100. Then, we use 'give him a pair of sunglasses' as the IN2N textual condition, iteratively editing the 3D scene and corresponding dataset. There are currently few personalized 3D scene editing methods. Therefore, we adopt 2D editing methods (e.g., AnyDoor~\citep{chen2023anydoor}) as another baseline for 3D scene editing. We use these methods to edit the 3D scene dataset and then train a model to obtain the edited 3D scene. When editing each image with AnyDoor, we keep the model's default parameters and turn on shape control.

\section{Ablation study}
\label{appen:ablation}

\begin{figure}[t]
  \centering
    \includegraphics[width=1.000\textwidth]{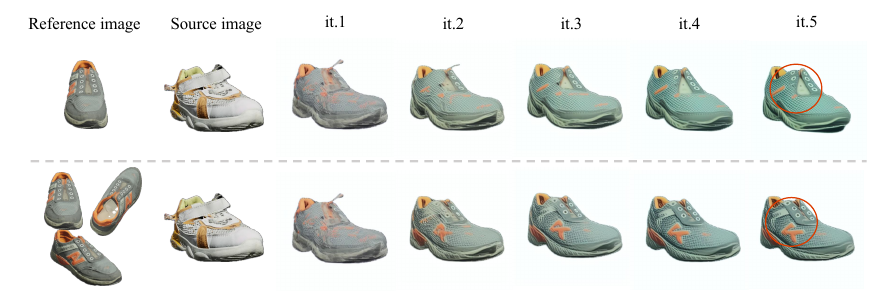}
  \caption{\textbf{Ablation study.} The top row depicts the insertion of features from a single reference image into the source image, along with the changes in the generated image at each iteration step. The bottom row illustrates the utilization of Feature Gradually Sampling to insert features from multiple reference images into the source image, as well as the changes in the generated image during each iteration. See Appendix~\ref{appen:gfs} for more details
  }
  \label{fig:gfs_fusion}
\end{figure}

{\bf Ablation on the components of FGS.}\label{appen:gfs} Feature Gradually Sampling strategy (See Section~\ref{sec:gfs}) is designed to address the issue where, in a single image scenario, insufficient subject information in the reference image may lead to the loss of certain structural details in the source image. As shown in Figure~\ref{fig:gfs_fusion} (top row), features highlighted within the red circle gradually weaken and eventually disappear during iterative layers (e.g. loss of the logo 'N'). Once these structures are lost, it becomes difficult to recover them in subsequent stages. FGS effectively mitigates this problem, as illustrated in Figure~\ref{fig:gfs_fusion} (bottom row), by preserving the structural details of the source image while injecting features from multiple reference images.

{\bf Controlling Subject Identity.} We can control at which step of denoising and which layer of the U-Net to start  \sname by setting \textit{S} and \textit{L}, respectively. Different settings of \textit{S} and \textit{L} parameters lead to different outcomes (See Figure~\ref{fig:shape_ctrl}). As \textit{S} and \textit{L} decrease, the number of iterations of \sname increases. This means that as more features from the reference image are injected into the source image, the generated result not only becomes visually more similar to the reference image but also structurally more alike. Conversely, the opposite is true.

\section{Evaluation details}\label{appen:evaluation}

\begin{figure}[!th]
  \centering
    \includegraphics[width=1.000\textwidth]{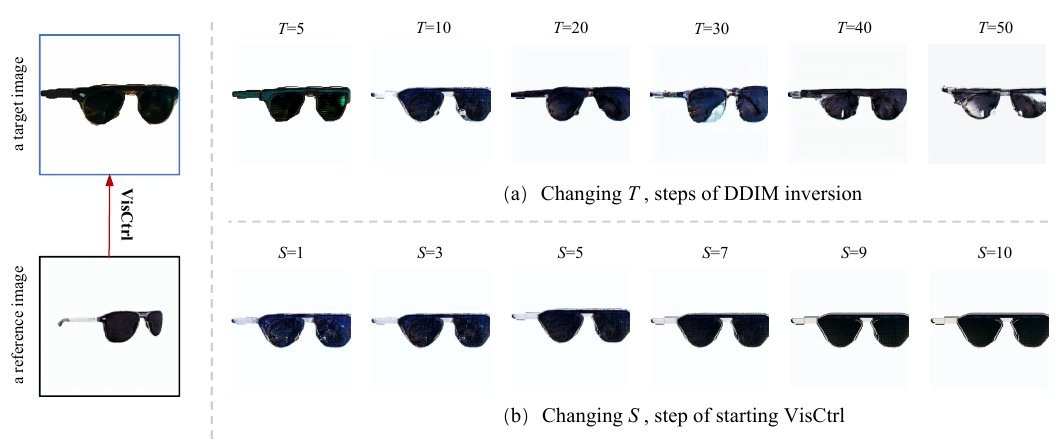}
  \caption{\textbf{Results at different denoising steps.} The top right row of the figure showcases the generated results with different denoising steps, while the bottom right row presents the generated results with different insertion steps when $T=10$.
  }
  \label{fig:diff_step}
\end{figure}

\subsection{Tasks}\label{appen:task}
We compared \sname with three other different methods, evaluating the editing results of four images (See Figure~\ref{fig:mutil_res}) and selecting three of these results for quantitative evaluations (See Table~\ref{tab:cmp_baselines}). Some input images are sourced from the DreamBooth dataset~\cite{ruiz2023dreambooth}, while others are obtained from the internet.

\subsection{Metrics}\label{appen:metrics}
For quantitative evaluations, we assess three criteria: (1) the adequacy of injected features from reference images, (2) the preservation of the source image's structure in the edited image, and (3) the consistency of background regions between images. We measure the fidelity of subjects between reference and generated images using CLIP-I~\cite{ruiz2023dreambooth}, which computes the average pairwise cosine similarity between CLIP~\cite{radford2021learning} embeddings of generated and real images. Additionally, we calculate the background LPIPS error (BG LPIPS) to quantify the preservation of background regions post-editing. This involves computing the LPIPS distance between background regions in the source and edited images, with background regions identified using the SAM object detector~\cite{kirillov2023segany}.
A lower BG LPIPS score indicates better preservation of the original image background. Finally, we employ the Structural Similarity Index Measure (SSIM) to gauge the similarity between the source image and the generated image, ensuring that the generated results maintain the overall structure of the source image.

In our work on video editing and 3D scene manipulation tasks, we employ the CLIP-I and LPIPS metrics. Additionally, we utilize CLIP Directional Similarity~\cite{gal2021stylegannada}, which quantifies the alignment between textual modifications and corresponding image alterations.


\begin{figure}[t]
  \centering
    \includegraphics[width=0.600\textwidth]{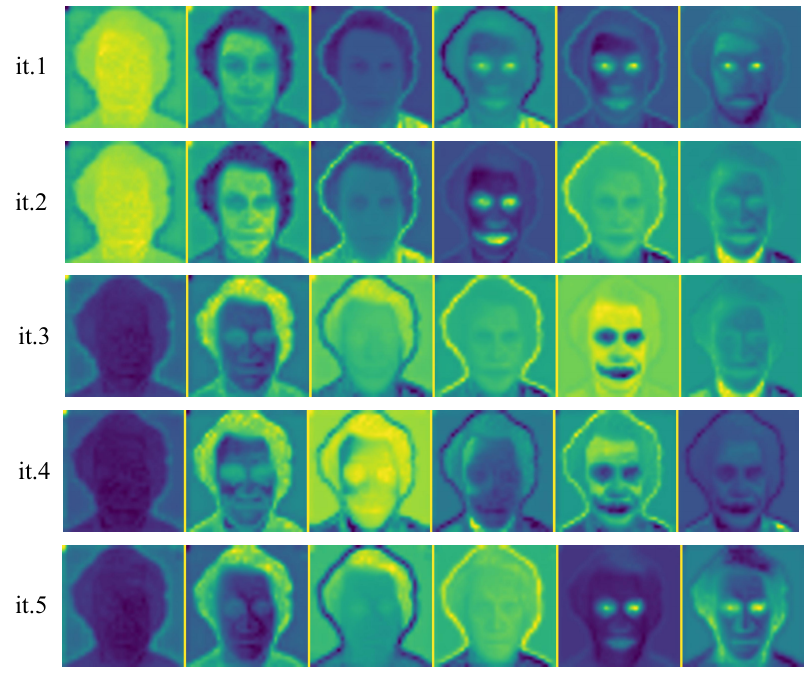}
  \caption{\textbf{Self-Attention maps under different iterations.} This representation reveals that the layout of the edited image is intrinsically embedded in the self-attention map from the initial iteration. At different stages of iteration, the attention map in the self-attention varies.
  }
  \label{fig:vis_self_attention}
\end{figure}

\section{Ethics Exploration}
\label{appd:ethics}

Similar to many AI technologies, text-to-image diffusion models may exhibit biases reflective of those inherent in the training data~\cite{sasha2023stable, perera2023analyzing}. Trained on extensive text and image datasets, these models might inadvertently learn and perpetuate biases, including stereotypes and prejudices, present within the data. For instance, if the training data contains skewed representations or descriptions of specific demographic groups, the model may produce biased images in response to related prompts.

\begin{figure}[!th]
  \centering
    \includegraphics[width=1.000\textwidth]{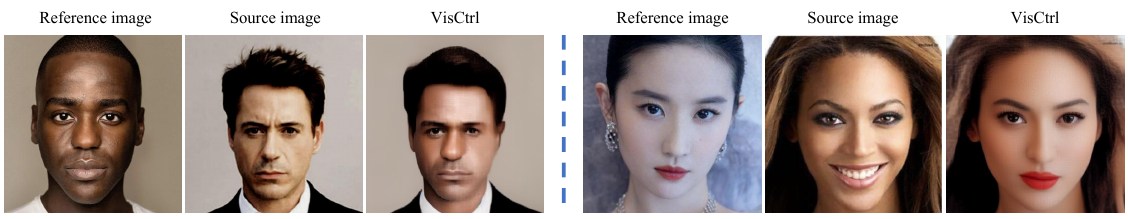}
  \caption{\textbf{Results on real human images across different races.} Evidently, the appearance features of the reference image can be seamlessly integrated into the source image, unaffected by skin color or gender.
  }
  \label{fig:ethics}
\end{figure}

However, \sname has been meticulously designed to mitigate bias within the text-to-image diffusion model generation process. It achieves this by first, not requiring retraining of the model and avoiding parameter updates; second, directly performing feature matching and injection in the latent space, thereby preventing bias introduction.

In Figure~\ref{fig:ethics}, we present our evaluation of facial feature injection across various skin tones and genders. It is crucial to note that significant disparities between the source and reference images tend to homogenize the skin color in the results. Consequently, we advocate for using \sname on subjects with similar racial backgrounds to achieve more satisfactory and authentic outcomes. Despite these potential disparities, the model ensures the preservation of most of the target subject's specific facial features, thereby reinforcing the credibility and accuracy of the final image.

\section{Failure Cases}
\label{append:failure_cases}

\begin{figure}[!ht]
    \centering
    \includegraphics[width=1.0\textwidth]{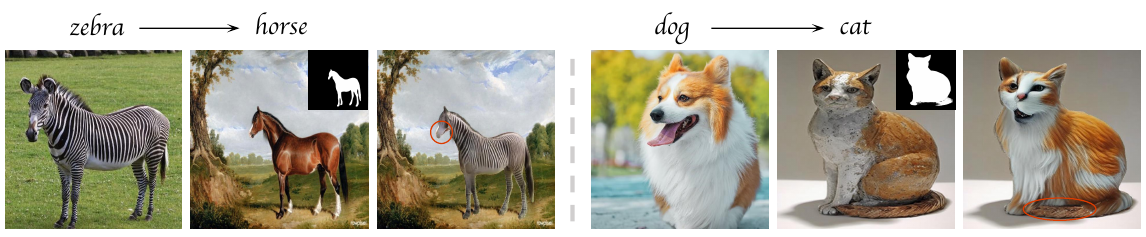}
    \caption{\textbf{Failure cases.} Our algorithm relies on SAM~\cite{kirillov2023segany} to obtain masks. Occasional inaccuracies in segmentation can result in errors in our generated results, as indicated by the red circles in the figure.}
    \label{fig:failure_cases}
\end{figure}

In this section, we highlight common failure cases. When intending to edit a specific subject within a source image, it is necessary to segment this subject using a segmentation algorithm. Subsequently, utilizing the reference image, \sname{} operations are performed to generate the desired subject. The final generated result is obtained by overlaying this generated subject with the corresponding mask. Consequently, if the mask produced by the segmentation algorithm is of poor quality, it may result in missing portions in the resulting image, as illustrated by the mouth of the horse and the tail of the cat in Figure~\ref{fig:failure_cases}.


\end{document}